\documentclass{article}
\usepackage{spconf,amsmath,graphicx,amsfonts}

\newcommand{\etal}{\textit{et al.}}

\usepackage{multirow}
\usepackage{verbatim}
\usepackage{arydshln}
\usepackage{caption}
\usepackage[labelformat=simple]{subcaption}

\usepackage{stfloats,enumitem}
\usepackage{hyperref}

\begin{document}
\twocolumn[{%
\vspace{30mm}
{ \large
\begin{itemize}[leftmargin=2.5cm, align=parleft, labelsep=2cm, itemsep=4ex,]

\item[\textbf{Citation}]{J. Lee and G. AlRegib, “Gradients as a Measure of Uncertainty in Neural Networks,” 27th \textit{IEEE International Conference on Image Processing (ICIP)}, Abu Dhabi, United Arab Emirates (UAE), 2020.}

\item[\textbf{Review}]{Date of Publication: October 25, 2020}


\item[\textbf{Bib}]  {@inproceedings\{lee2020gradients,\\
    title=\{Gradients as a Measure of Uncertainty in Neural Networks\},\\
    author=\{Lee, Jinsol and AlRegib, Ghassan\},\\
    booktitle=\{IEEE International Conference on Image Processing (ICIP)\},\\
    year=\{2020\}\}}

\item[\textbf{Copyright}]{\textcopyright 2020 IEEE. Personal use of this material is permitted. Permission from IEEE must be obtained for all other uses, in any current or future media, including reprinting/republishing this material for advertising or promotional purposes,
creating new collective works, for resale or redistribution to servers or lists, or reuse of any copyrighted component
of this work in other works.}

\item[\textbf{Contact}]{
\{jinsol.lee, alregib\}@gatech.edu\\
\url{https://ghassanalregib.info/}\\}
\end{itemize}
}}]
\newpage
\clearpage
\setcounter{page}{1}


\title{Gradients as a measure of uncertainty in neural networks}
%
\name{Jinsol Lee and Ghassan AlRegib}
\address{OLIVES at the Center for Signal and Information Processing,\\ School of Electrical and Computer Engineering,\\ Georgia Institute of Technology, Atlanta, GA, 30332-0250\\ \{jinsol.lee, alregib\}@gatech.edu}

\ninept
\maketitle
\begin{abstract}
Despite tremendous success of modern neural networks, they are known to be overconfident even when the model encounters inputs with unfamiliar conditions. Detecting such inputs is vital to preventing models from making naive predictions that may jeopardize real-world applications of neural networks. In this paper, we address the challenging problem of devising a simple yet effective measure of uncertainty in deep neural networks. Specifically, we propose to utilize backpropagated gradients to quantify the uncertainty of trained models. Gradients depict the required amount of change for a model to properly represent given inputs, thus providing a valuable insight into how familiar and certain the model is regarding the inputs. We demonstrate the effectiveness of gradients as a measure of model uncertainty in applications of detecting unfamiliar inputs, including out-of-distribution and corrupted samples. We show that our gradient-based method outperforms state-of-the-art methods by up to 4.8\% of AUROC score in out-of-distribution detection and 35.7\% in corrupted input detection.

\end{abstract}
%

\section{Introduction}

Deep neural networks (DNNs) have achieved significant improvements in a variety of applications, including object detection~\cite{ren2015fasterrcnn}, image classification~\cite{he2016resnet}, and natural language processing~\cite{tenney2019bert}. Despite these advancements, DNNs are prone to failure when deployed in real-world environments as they often encounter data that diverges from training conditions~\cite{temel2018cureor, temel2019multifarious}. Neural networks make predictions regardless of their familiarity with given inputs. Combined with their overconfident nature~\cite{goodfellow2014adversarial,guo2017calibration}, the difficulty in distinguishing naive predictions about unfamiliar inputs from knowledgeable predictions about familiar inputs hinders practical implementation of DNNs. To ensure a certain level of reliable performance, models must be capable of identifying unfamiliar inputs, which then can be handled via predetermined fallback mechanisms instead of making erroneous conjectures. 

At the core of detecting unfamiliar inputs lies the concept of uncertainty and the methodology to measure it. In this work, we mainly consider \textit{model uncertainty}, which refers to the uncertainty in model parameters due to limited data and knowledge~\cite{kendall2017uncertainties}. It is directly related to the problem of out-of-distribution (OOD) detection, where the goal is to detect the samples drawn from different data distribution than in-distribution or training samples. Recent studies~\cite{hendrycks2016baseline,liang2017odin} utilized the prediction probability statistics as an uncertainty measure. Hendrycks \& Gimpel~\cite{hendrycks2016baseline} introduced a baseline method of thresholding samples based on the predicted softmax probability. Liang~\etal~\cite{liang2017odin} improved their method with input and output processing to further distinguish in-distribution and OOD samples. DeVries \& Taylor~\cite{devries2018confidence} utilized prediction confidence as an uncertainty measure, obtained from an augmented confidence estimation branch on a pre-trained classifier. Lee~\etal~\cite{lee2018mahalanobis} proposed a confidence metric with Mahalanobis distance to characterize OOD samples. These approaches investigate model uncertainty with activation-based measures and typically utilize additional modules for pre- and post-processing as well as quantifying uncertainty that require calibrations.

In this work, we propose a simple yet effective method to measure model uncertainty of deep neural networks. Specifically, \textit{we utilize backpropagated gradients to probe the knowledge of fully trained networks and determine the model's familiarity with given inputs}. Our intuition is that gradients correspond to the amount of changes it requires to properly represent a given sample. In comparison with aforementioned activation-based methods~\cite{hendrycks2016baseline,liang2017odin,devries2018confidence,lee2018mahalanobis}, our method in estimating model uncertainty takes into account both activation and loss associated with given inputs as well as the relationship between the two, preserving more information about model uncertainty. In addition, our method does not require any calibration or extra processing. We empirically show that our gradient-based method outperforms activation-based methods across multiple tasks and datasets. The contributions of this paper are three-fold.

\begin{itemize}[leftmargin=4mm,itemsep=-1mm]
\vspace{-5mm}
    \item We provide an interpretation of gradients in the space of models from a perspective of model uncertainty.
    \item We introduce an efficient framework to generate gradients for uncertainty characterization. 
    \item We apply our gradient-based approach to applications of out-of-distribution detection and corrupted input detection to achieve the state-of-the-art performance.
\end{itemize}

\section{Background}\label{sec:background}

\begin{figure}[t]
    \centering
    \includegraphics[width=0.8\linewidth, trim=6.9cm 5.7cm 3.9cm 6cm, clip]{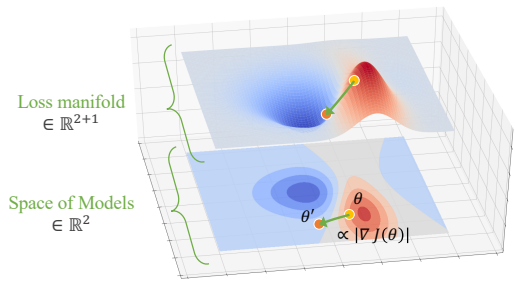}
    \caption{Illustration of a space of models.}
    \label{fig:space_of_model}
    \vspace{-5mm}
\end{figure}

Given any single network architecture, the set of models is defined by all possible configurations of its parameters. In the immense space of these models, we manage to find a seemingly endless number of parameterizations that perform well via various initialization and optimization techniques. In particular, stochastic gradient-based optimization has been fundamental to the success of modern neural networks. It is built on the assumption that the model is defined on a relatively smooth space where the first derivatives exist~\cite{allen2018overparameterization}. Each update of gradient-based optimization, as shown in Eq.~(\ref{eq:gradient_optimization}), aims to minimize an objective function $J(\theta)$ with respect to model parameters $\theta \in \mathbb{R}^d$ by updating the parameters in the opposite direction of the gradients of the objective function $\nabla_\theta J(\theta)$. 
\begin{equation} \label{eq:gradient_optimization}
    \theta' = \theta - \eta \cdot \nabla_\theta J(\theta)
\end{equation}

The direction of the gradients determines the \emph{direction} of the update, and the magnitude of gradients scaled by learning rate $\eta$ governs the \emph{size} of the update. It is a process of maneuvering the space of models from one parameterization to another. In Fig.~\ref{fig:space_of_model}, we visualize a space of models for an architecture of two parameters and its arbitrary loss manifold. The model space is two-dimensional as defined by the number of parameters. The loss manifold is depicted with one extra dimension since, for each parameterization, a loss value can be computed with respect to a given input. In training, each iteration involves finding a model parameterization by optimizing on the loss manifold. On the space of models, a model that performs better than the current parameterization lies in the opposite direction of the gradients at a distance of some proportion of the gradient magnitude.
 
\section{Gradients as a measure of uncertainty}

\begin{figure*}[hb]
\vspace{-2.5mm}
    \centering
    \begin{subfigure}[t]{.4\textwidth}
        \centering
        \includegraphics[width=\linewidth, trim=6cm 4.6cm 6cm 4.5cm, clip]{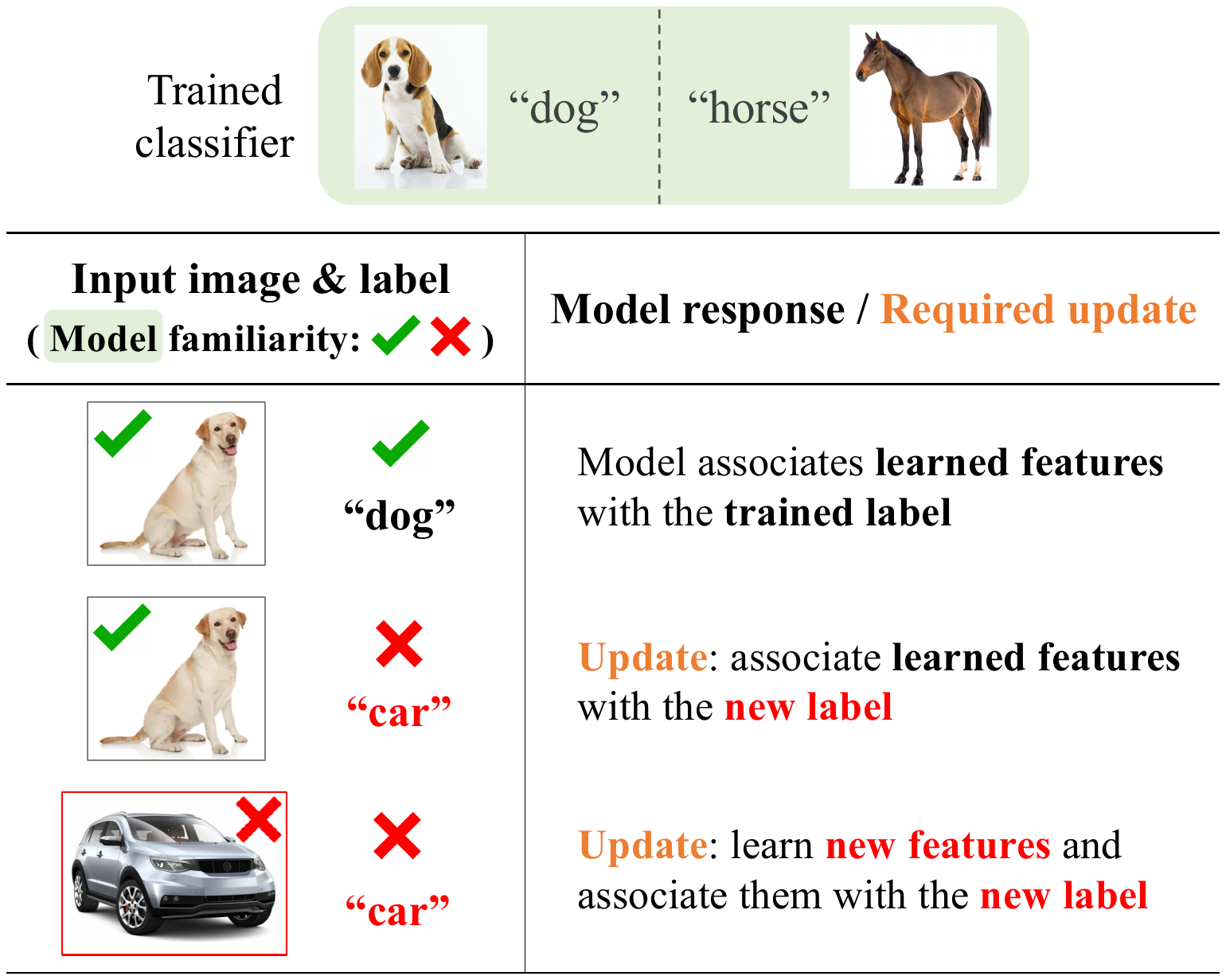}
        \subcaption{The effect of a confounding label.}
        \label{sfig:confounding_label}
    \end{subfigure}\hfill
    \begin{subfigure}[t]{.58\textwidth}
        \includegraphics[width=\linewidth, trim=4.5cm 6.1cm 4.3cm 5.7cm, clip]{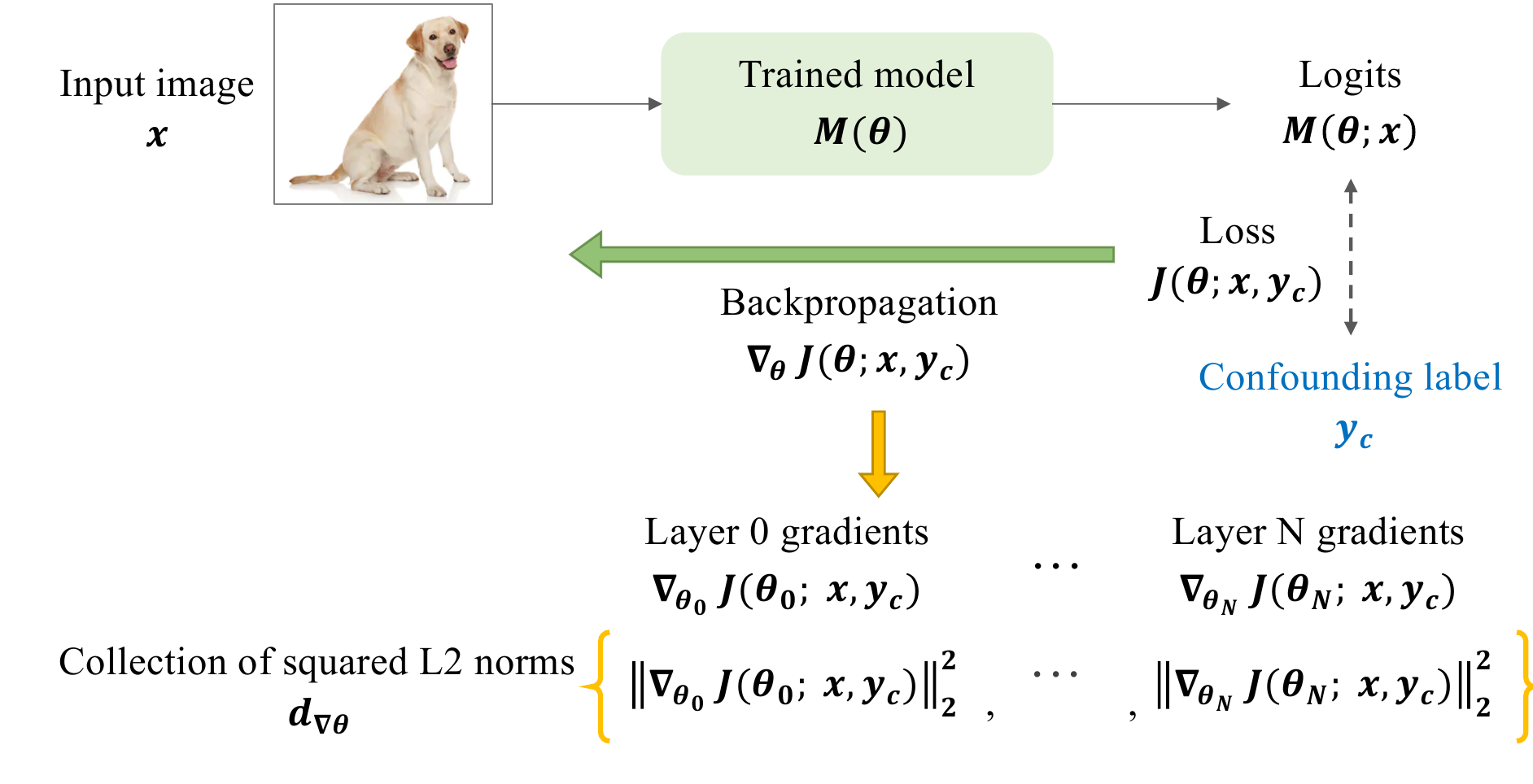}
        \subcaption{Gradient generation framework with confounding label.}
        \label{sfig:gradient_generation}
    \end{subfigure}
\vspace{-1.5mm}
\caption{Quantifying uncertainty of neural networks with gradients. Confounding labels are utilized for effective gradient generation.}
\label{fig:gradients}
\vspace{-3mm}
\end{figure*}

Gradient-based optimization involves larger updates when there is a larger gap between predictions and correct labels for given inputs. It implies that the model requires more significant adjustments to its parameters, as it has not learned enough features to represent the inputs or relationships between learned features and classes for correct prediction. We propose to utilize gradients to quantify uncertainty of fully trained neural networks. Our intuition is that the magnitude of the gradients is directly proportional to the amount of each update in training. In the context of the space of models, the size of each update represents the distance between two distinct model parameterizations: one before the update ($\theta$) and the other afterward ($\theta'$), as shown in Fig.~\ref{fig:space_of_model}. We can exploit this idea of distance in the space of models, represented by the magnitude of gradients, as a measure of uncertainty. \textit{During inference}, gradients can be generated in a similar manner as in training process. \textit{The obtained gradients are not pertinent to model updates; however, they contain more reliable information about the familiarity of the model with given inputs, as the model in question is a converged solution and the required model update should not be drastic for inputs that are similar to training data.} In the following sections, we describe the process of generating and utilizing such gradients in detail.

\subsection{Gradient generation framework}\label{ssec:gradients_generation}

To use gradients as a measure of uncertainty for a fully trained network, we first address the process of obtaining gradients. It requires backpropagation of arbitrary loss between predictions that the model made for given inputs and their ground truth labels; however, the labels are not available at inference. While we may be able to analyze gradients for every class that the model is trained for, the number of loss and gradient computations increases linearly with the number of possible classes for the model. For a more efficient gradient generation process, we construct confounding labels. We define a \textit{confounding label} as a label that is different from ordinary labels on which a model is trained. In an image classification setting, an ordinary label consists of a single class (i.e. one-hot vector), whereas a confounding label may include multiple classes or none---a vector of length $C$ with $n$ number of $1$'s where $C$ is the number of classes in training and $n \in \{0, \dots, C\} \setminus \{1\}$. We describe the effect of a confounding label in Fig.~\ref{sfig:confounding_label}. Consider a trained classifier that is trained with images and labels of two classes: ``dog" and ``horse". With an ordinary label (i.e. ``dog" or ``horse"), when a model is familiar with an input image, the model has features learned during training to represent the input. In addition, the model is able to associate the already-learned features with one of the trained classes provided by the given ordinary label. On the other hand, a confounding label is specifically designed to be unfamiliar to the model. With a confounding label (i.e. ``car") and a familiar input, the model does not need to learn features again. The only necessary adjustments to model parameters would be to learn the relationship between the already-learned features and the confounding label. If the model is unfamiliar with the given input, however, the amount of updates required to 1) learn new features to properly represent the new input, and 2) associate the new features with the confounding label, would be larger.

We discuss the framework to collect gradients with confounding labels from a fully trained network in Fig.~\ref{sfig:gradient_generation}. Given an input image, the model yields logits. We utilize binary cross entropy loss between the logits and a confounding label as shown in Eq.~(\ref{eq:bce}), where $\hat{y_i}$ is the predicted probability for class $i$ and $y_i$ is the true probability represented by the confounding label. Through backpropagation of the loss, gradients are generated for each set of model parameters (i.e. weight and bias parameters of network layers). While any form of conserving the magnitude would be valid, we measure the squared $L_{2}$ distance of the gradients for each parameter set and concatenate them to represent the given input.
\begin{equation} \label{eq:bce}
    J(\theta)= \frac{1}{C} \sum^{C}_{i=0} \left(y_i \cdot \log\left(\hat{y_i}\right) + \left(1 - y_i\right)\cdot\log\left(1-\hat{y_i}\right)\right)
    \vspace{-2mm}
\end{equation}

\subsection{Demonstration of the effectiveness of gradients}

\begin{figure*}[t]
\vspace{-2mm}
    \centering
    \begin{subfigure}[t]{.68\textwidth}
        \includegraphics[height=4.9cm]{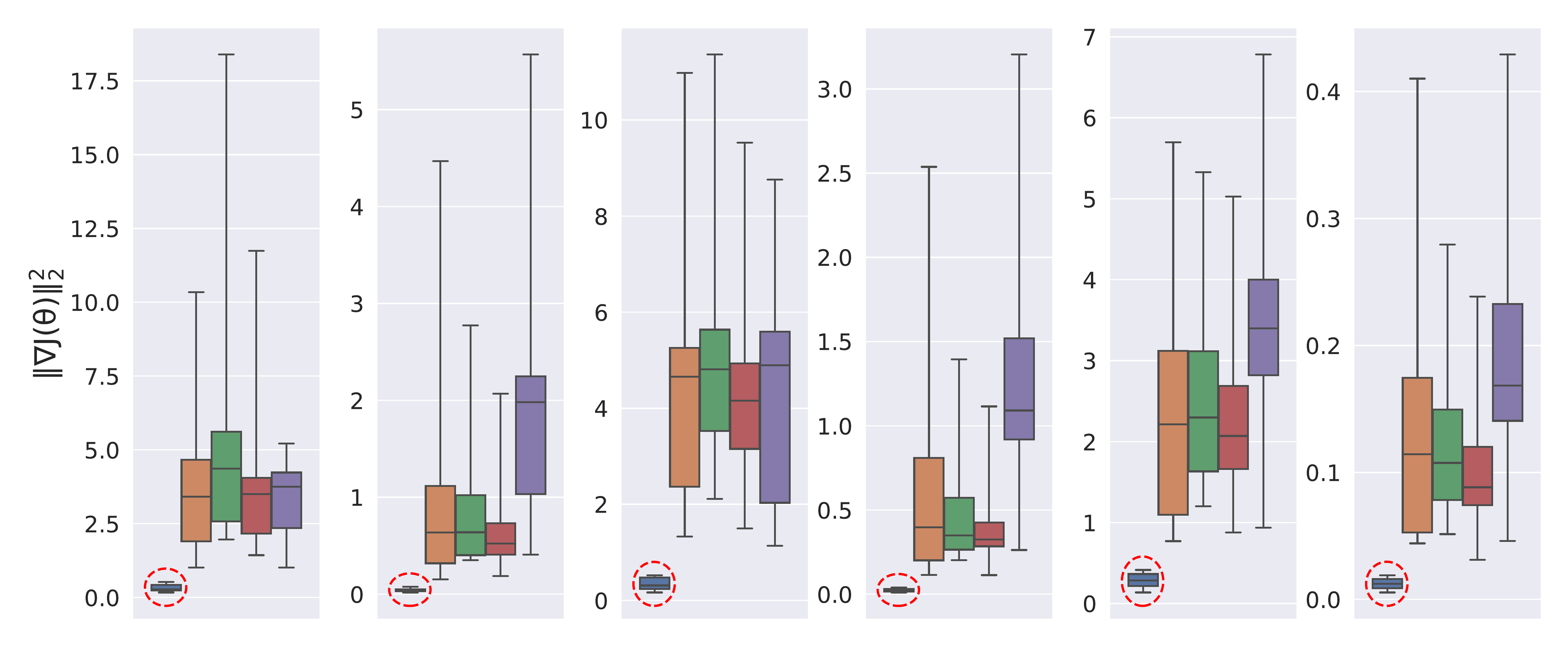}
        \subcaption{Squared $L_2$ norms of gradients at different parameter sets: $\Vert \nabla J (\theta) \Vert_2^2$.}
        \label{sfig:gradient_norm}
    \end{subfigure}
    \begin{subfigure}[t]{.17\textwidth}
        \includegraphics[height=4.9cm]{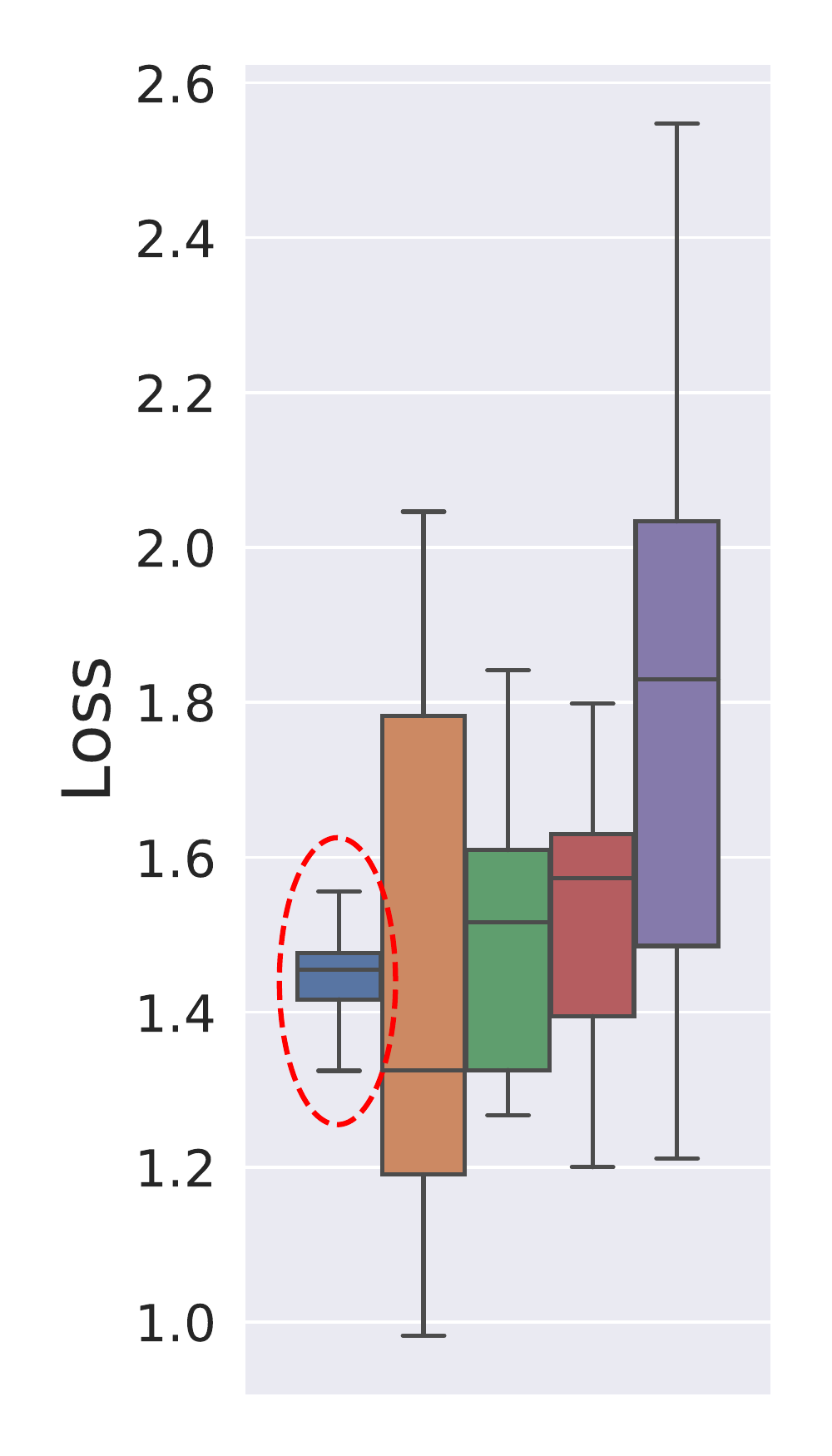}
        \subcaption{Loss: $J (\theta)$}
        \label{sfig:loss}
    \end{subfigure}
    \begin{subfigure}[b]{.14\textwidth}
        \includegraphics[width=\textwidth]{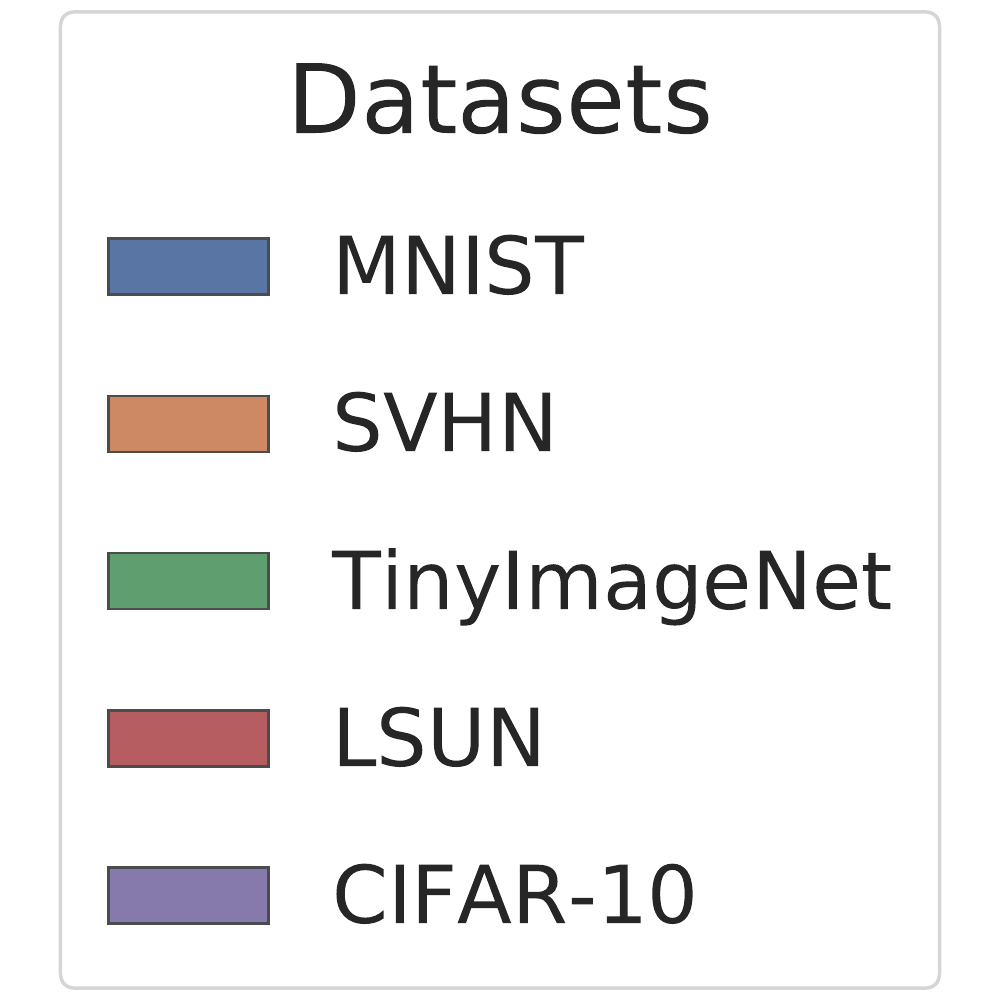}\vspace{8mm}
    \end{subfigure}\hfill
\vspace{-1mm}
\caption{Comparison between gradients and loss in distinguishing in-distribution and out-of-distribution datasets. Each plot shows distributions of per-class average values of considered attributes ($\Vert \nabla J (\theta) \Vert_2^2$ or $J (\theta)$) for all datasets.}
\vspace{-2mm}
\label{fig:toy_example}
\end{figure*}

To demonstrate the effectiveness of gradients as a measure of model uncertainty, we present preliminary experiments in out-of-distribution detection setups. First, we train a ResNet-18~\cite{he2016resnet} model with the training set of MNIST~\cite{lecun1998mnist}. The trained model is then used to generate gradients as described in Section~\ref{ssec:gradients_generation} on the test sets of MNIST, CIFAR-10~\cite{krizhevsky2009cifar}, TinyImageNet~\cite{deng2009imagenet}, LSUN~\cite{yu2015lsun}, and SVHN~\cite{netzer2011svhn}. MNIST is chosen to be an in-distribution dataset as it is the simplest among all considered datasets. We employ a network architecture that is large enough to learn classification for all datasets. With the magnitude of gradients, we show the disparity between the learned features in training and the necessary features to represent test images. 

In Fig.~\ref{sfig:gradient_norm}, we visualize the distributions of per-class average magnitude of gradients, captured in different parameter sets of the model. Note that the distribution of MNIST is highlighted in red circle in each plot for clarity. The separation between in-distribution and OOD datasets is more evident in some parts of the network than others because each layer captures information about different aspects of given inputs. But overall, we observe a sharp distinction based on the familiarity of the model, with smaller gradient magnitude for in-distribution/familiar datasets and significantly larger for OOD/unfamiliar datasets. In addition, we show the distribution of loss values in Fig.~\ref{sfig:loss} for further analysis. Gradients and loss are intertwined in the process of backpropagation, but the loss is limited to a single value per iteration. We observe apparent overlaps of loss values across different datasets, demonstrating that loss alone is inadequate for detecting unfamiliar inputs. On the contrary, the gradients are of the same dimensions as their corresponding parameter sets, preserving more information about the current state of the model and the necessary adjustment for better representation of given inputs. 
 
\section{Experiments}

We validate the effectiveness of our proposed method in applications that deal with some form of unfamiliar inputs, including out-of-distribution detection and corrupted input detection. For each application, we distinguish the familiarity of datasets based on the model training process---familiar if the model has been exposed to the dataset in training and unfamiliar otherwise. Specifically, we train ResNet with 18 layers on training set of a dataset and obtain gradient-based representations from test sets of both familiar and unfamiliar datasets. To generate gradients, we utilize a confounding label where all classes are positive (i.e. a vector of all $1$'s). For each set of familiar and unfamiliar datasets, we use a 40\%-40\%-20\% train-validation-test split of the collected gradient representations and train a simple binary detector of 2 fully-connected layers as an unfamiliar input detector.

\subsection{Out-of-distribution detection}\label{ssec:ood}

For out-of-distribution (OOD) detection experiments, we use various image classification datasets: CIFAR-10 and SVHN as in-distribution, additional TinyImageNet and LSUN as OOD. For fair comparison, we utilize the same experiment setups to replicate results from other state-of-the-art OOD methods~\cite{hendrycks2016baseline, liang2017odin, lee2018mahalanobis}. For evaluation, we measure the following metrics: detection accuracy, area under receiver operating characteristic curve (AUROC), and area under precision-recall curve (AUPR). The results of OOD detection are reported in Table~\ref{table:ood} and the method with the best result for each metric is highlighted. While both ODIN~\cite{liang2017odin} and Mahalanobis~\cite{lee2018mahalanobis} methods involve input and output processing, we report vanilla results (i.e. without any processing) for Mahalanobis method as it shows better performance than ODIN when the additional procedures are allowed. Note that these procedures require delicate calibration of hyperparameters, while our method does not require any. We observe that our method outperforms all other techniques when CIFAR-10 is used as in-distribution and SVHN as OOD, and SVHN as in-distribution and all others as OOD, up to 9\% increase in performance. Our method is especially effective when there is a larger difference in the complexity of in-distribution and OOD datasets. SVHN comprises color images of house numbers ranging from 0 to 9, while the other datasets include more complex natural images of diverse classes. Our method still shows very comparable result when the model is trained on CIFAR-10 and LSUN is used as OOD. We remark that our gradient-based method outperforms all activation-based methods in most cases. \vspace{-1mm}

\begin{table*}[!htb]
\captionsetup{width=.92\linewidth}
\caption{The results of distinguishing in- and out-of-distribution test set data on OOD benchmarks. All values are in percentages and the best results are highlighted in bold. For Mahalanobis~\cite{lee2018mahalanobis} method, V indicates vanilla results without input pre-processing or feature ensemble, and P+FE includes both.}
\centering
\resizebox{0.97\linewidth}{!}{
\renewcommand{\arraystretch}{1.5}
\begin{tabular}{cc|ccc}
\hline \hline
\multicolumn{2}{c|} {Dataset Distribution} & Detection Accuracy & AUROC & AUPR \\ \hdashline
In & Out & \multicolumn{3}{c}{Baseline~\cite{hendrycks2016baseline} / ODIN~\cite{liang2017odin} / Mahalanobis (V)~\cite{lee2018mahalanobis} / Mahalanobis (P+FE)~\cite{lee2018mahalanobis} / Ours} \\ \hline \hline
\multirow{3}{*}{CIFAR-10} & SVHN &  83.36 / 88.81 / 79.39 / 91.95 / \textbf{98.04}  &	88.30 / 94.93 / 85.03 / 97.10 / \textbf{99.84} &	88.26 / 95.45 / 86.15 / 96.12 / \textbf{99.98}\\ \cline{2-5} 
 & TinyImageNet &  84.01 / 85.21 / 83.60 / \textbf{97.45} / 86.17 &	90.06 / 91.86 / 88.93 / \textbf{99.68} / 93.18 &	89.26 / 91.60 / 88.59 / \textbf{99.60} / 92.66 \\ \cline{2-5} 
 & LSUN &  87.34 / 88.42 / 85.02 / \textbf{98.60} / 98.37 &	92.79 / 94.48 / 90.11 / \textbf{99.86} / \textbf{99.86} &	92.30 / 94.22 / 89.80 / 99.82 / \textbf{99.87} \\ \hline
\multirow{3}{*}{SVHN} & CIFAR-10 &  79.98 / 80.12 / 74.10 / 88.84 / \textbf{97.90} &	81.50 / 81.49 / 79.31 / 95.05 / \textbf{99.79} &	81.01 / 80.95 / 80.83 / 90.25 / \textbf{98.11} \\ \cline{2-5} 
 & TinyImageNet &  81.70 / 81.92 / 79.35 / 96.17 / \textbf{97.74} & 83.69 / 83.82 / 83.85 / 99.23 / \textbf{99.77}	& 82.54 / 82.60 / 85.50 / \textbf{98.17} / 97.93 \\ \cline{2-5} 
 & LSUN &  80.96 / 81.15 / 79.52 / 97.50 / \textbf{99.04} &	82.85 / 82.98 / 83.02 / 99.54 / \textbf{99.93} &	81.97 / 82.01 / 84.67 / 98.84 / \textbf{99.21} \\ \hline \hline
\end{tabular}}
\label{table:ood}
\vspace{-1mm}
\end{table*}

\vspace{-1mm}
\subsection{Corrupted input detection}

In addition to model uncertainty, we consider \textit{aleatoric uncertainty}, another type of uncertainty that is related to noise in observations~\cite{kendall2017uncertainties}. Some examples of the cause of this uncertainty are imperfect data acquisition and environmental factors such as motion blur or weather conditions. Aleatoric uncertainty is considered irreducible even with more data due to the inherent nature of noise, so most approaches focus on designing a robust system that can handle both pristine and noisy inputs. We apply our gradient-based uncertainty measure to directly detect corrupted inputs. We utilize the Mahalanobis method from out-of-distribution detection as a comparison, as corrupted inputs can also be considered as OOD examples and the Mahalanobis method shows the best performance among all compared methods. 

For corrupted input detection experiments, we use image classification datasets that are designed for benchmarking robustness of neural networks with realistic challenging conditions. CIFAR-10-C~\cite{hendrycks2019robustness} consists of 19 diverse corruption types of 4 categories, including noise, blur, weather and digital, at 5 different severity levels that are applied to validation images of CIFAR-10 dataset. CURE-TSR~\cite{temel2017curetsr} is a traffic sign recognition dataset that includes real-world and simulated challenging conditions of 11 types and 5 severity levels. For each dataset, a model is trained on training images that are free of corruptions, and the gradients are collected for test sets of both pristine and corrupted images. Among all image corruption types from both datasets, we select common or similar ones to report AUROC scores in Table~\ref{table:corrupted}. For CIFAR-10-C dataset, our method outperforms the Mahalanobis method for all corruption types and levels. It shows saturated performance even from severity level 1, while the Mahalanobis method generally shows an increase at higher levels of corruption. Regarding CURE-TSR dataset, our method still outperforms in most cases and show comparable results otherwise, but both methods show more variations in performance across severity levels of corruptions. Note the gap in AUROC scores between the methods at lower levels of corruptions---this implies that our gradient-based method can characterize corruptions effectively, compared to the activation-based method, even at a subtle degree.

\begin{table}[!ht]
\captionsetup{width=.95\linewidth}
\caption{The results of distinguishing pristine and corrupted images. All values are AUROC scores in percentages and the best results are highlighted in bold.}
\resizebox{\linewidth}{!}{
\renewcommand{\arraystretch}{1.8}
\begin{tabular}{@{\hspace{0.7ex}}c|c|c@{\hspace{2ex}}c@{\hspace{2ex}}c@{\hspace{2ex}}c@{\hspace{2ex}}c}
\hline\hline
\multirow{2}{*}{\rotatebox[origin=c]{90}{\small Dataset}} & \small Method & \multicolumn{5}{c}{Mahalanobis~\cite{lee2018mahalanobis} / Ours} \\ \cdashline{2-7}
 & \small Corruption & Level 1 & Level 2 & Level 3 & Level 4 & Level 5 \\ \hline\hline
\multirow{8}{*}{\rotatebox[origin=c]{90}{\small CIFAR-10-C}} 
 & \small Noise & 96.63 / \textbf{99.95} & 98.73 / \textbf{99.97} & 99.46 / \textbf{99.99} & 99.62 / \textbf{99.97} & 99.71 / \textbf{99.99} \\  
 & \small LensBlur & 94.22 / \textbf{99.95} & 97.51 / \textbf{99.99} & 99.26 / \textbf{100.0} & 99.78 / \textbf{100.0} & 99.89 / \textbf{100.0} \\  
 & \footnotesize GaussianBlur & 94.19 / \textbf{99.94} & 99.28 / \textbf{100.0} & 99.76 / \textbf{100.0} & 99.86 / \textbf{100.0} & 99.80 / \textbf{100.0} \\  
 & \small DirtyLens & 93.37 / \textbf{99.94} & 95.31 / \textbf{99.93} & 95.66 / \textbf{99.96} & 95.37 / \textbf{99.92} & 97.43 / \textbf{99.96} \\  
 & \small Exposure & 91.39 / \textbf{99.87} & 91.00 / \textbf{99.85} & 90.71 / \textbf{99.88} & 90.58 / \textbf{99.85} & 90.68 / \textbf{99.87} \\  
 & \small Snow & 93.64 / \textbf{99.94} & 96.50 / \textbf{99.94} & 94.44 / \textbf{99.95} & 94.22 / \textbf{99.95} & 95.25 / \textbf{99.92} \\  
 & \small Haze & 95.52 / \textbf{99.95} & 98.35 / \textbf{99.99} & 99.28 / \textbf{100.0} & 99.71 / \textbf{99.99} & 99.94 / \textbf{100.0} \\  
 & \small Decolor & 93.51 / \textbf{99.96} & 93.55 / \textbf{99.96} & 90.30 / \textbf{99.82} & 89.86 / \textbf{99.75} & 90.43 / \textbf{99.83} \\ \hdashline
 \multirow{8}{*}{\rotatebox[origin=c]{90}{\small CURE-TSR}} 
 & \small Noise & 25.46 / \textbf{50.20} & 47.54 / \textbf{63.87} & 47.32 / \textbf{81.20} & 66.19 / \textbf{91.16} & 83.14 / \textbf{94.81} \\  
 & \small LensBlur & 48.06 / \textbf{72.63} & 71.61 / \textbf{87.58} & 86.59 / \textbf{92.56} & 92.19 / \textbf{93.90} & 94.90 / \textbf{95.65} \\  
 & \footnotesize GaussianBlur & 66.44 / \textbf{83.07} & 77.67 / \textbf{86.94} & 93.15 / \textbf{94.35} & 80.78 / \textbf{94.51} & \textbf{97.36} / 96.53 \\  
 & \small DirtyLens & 29.78 / \textbf{51.21} & 29.28 / \textbf{59.10} & 46.60 / \textbf{82.10} & 73.36 / \textbf{91.87} & 98.50 / \textbf{98.70} \\  
 & \small Exposure & 74.90 / \textbf{88.13} & \textbf{99.96} / 96.78 & \textbf{99.99} / 99.26 & \textbf{100.0} / 99.80 & \textbf{100.0} / 99.90 \\  
 & \small Snow & 28.11 / \textbf{61.34} & 61.28 / \textbf{80.52} & 89.89 / \textbf{91.30} & \textbf{99.34} / 96.13 & \textbf{99.98} / 97.66 \\  
 & \small Haze & 66.51 / \textbf{95.83} & 97.86 / \textbf{99.50} & \textbf{100.0} / 99.95 & \textbf{100.0} / 99.87 & \textbf{100.0} / 99.88 \\  
 & \small Decolor & 48.37 / \textbf{62.36} & 60.55 / \textbf{81.30} & 71.73 / \textbf{89.93} & 87.29 / \textbf{95.42} & 89.68 / \textbf{96.91} \\ \hline\hline
\end{tabular}}
\label{table:corrupted}
\end{table}

\section{Conclusion}

In this paper, we proposed a simple yet effective measure of uncertainty in deep neural networks with backpropagated gradients. We introduced the concept of the space of models in terms of optimization and motivated the utility of gradients as a distance measure in the space of model that corresponds to the required amount of adjustment to model parameters to properly represent given inputs. We designed a framework for efficient generation of gradients with confounding labels, and utilized the gradients to quantify uncertainty of fully trained networks regarding the given inputs. We validated our approach in the applications of detecting unfamiliar inputs, including out-of-distribution samples and corrupted images. Our gradient-based method outperforms activation-based state-of-the-art methods by up to 4.8\% of AUROC score in out-of-distribution detection and 35.7\% in corrupted input detection.
\newpage

\bibliographystyle{IEEEbib}
\bibliography{bib}

\end{document}